\documentclass{ifacconf}

\usepackage{graphicx}      
\usepackage{natbib}        
\usepackage{amsmath}
\usepackage{amsfonts}
\usepackage{caption}
\usepackage{subcaption}
\usepackage{svg}
\usepackage{listings}
\usepackage{stfloats}
\usepackage{float}
\begin{document}
\begin{frontmatter}
\title{Semantic and Topological Mapping using Intersection Identification}


\author{Scott Fredriksson, Akshit Saradagi and George Nikolakopoulos} 

\address{Robotics and Artificial Intelligence Team\\Luleå University of Technology, 
   Luleå, Sweden\\ (e-mail: scofre@ltu.se).}

\begin{abstract}                
This article presents a novel approach to identifying and classifying intersections for semantic and topological mapping. More specifically, the proposed novel approach has the merit of generating a semantically meaningful map containing intersections, pathways, dead ends, and pathways leading to unexplored frontiers. Furthermore, the resulting semantic map can be used to generate a sparse topological map representation, that can be utilized by robots for global navigation. The proposed solution also introduces a built-in filtering to handle noises in the environment, to remove openings in the map that the robot cannot pass, and to remove small objects to optimize and simplify the overall mapping results. The efficacy of the proposed semantic and topological mapping method is demonstrated over a map of an indoor structured environment that is built from experimental data. The proposed framework, when compared with similar state-of-the-art topological mapping solutions, is able to produce a map with up to 89\% fewer nodes than the next best solution. 
\end{abstract}


\end{frontmatter}

\section{Introduction and Related Work}

\label{sec:Introduction}
Navigating in known or unknown environments is one of the most challenging problems in robotics and a popular approach is to utilize grid-based global navigation algorithms, such as the A* algorithm (\cite{aStar}). Such approaches grow computationally heavy, especially when navigating long distances or when having to handle a complex map (\cite{aStarPreformence}). A solution to overcome this problem is to utilize a topological map. 
Topological mapping aims to represent the environment as a set of nodes and their interconnections, often as a skeleton map of the environment (\cite{topologicalDef}). Thus, the topological map is a high-level, sparser and more semantically meaningful representation of the environment.

Topological mapping has multiple uses in both online and offline scenarios. In offline situations, the topological map can be used to generate global robot paths in known environments, while in online exploration scenarios, a robot can extract semantic information from the topological map to perceive its environment, e.g., if it is approaching an intersection or a dead end. There is also a use for topological mapping in multi-agent robotics, mainly due to its sparse nature, which requires less bandwidth to send a topological map than a complete map between agents. 

One of the most common approaches for creating topological maps is to use a Voronoi graph (\cite{voronoi}). Voronoi maps contain a set of points with equal distances to two or more obstacles and such points result in a topological map represented as a skeleton map of the environment, where each branch and endpoint represents a node in the map. A Voronoi map is generally constructed using range data from a LiDAR but can also be approximated from an occupancy grid (\cite{5650794}). \cite{doi:10.1177/02783640022066770} extended the idea of the General Voronoi Diagram (GVD) to General Voronoi Graph (GVG) and used it to solve local navigation for robots. \cite{928558} later proposed a solution for topological mapping using GVG, while \cite{1570793} presented an Extended Voronoi Graph (EVG) with improvements over GVG with additional rules to improve its behavior in big open rooms. The issue with Voronoi-based topological maps is that they are susceptible to noise from the environment and data collection and tend to create maps with many nodes compared to other solutions. An outcome of this drawback is that Voronoi graphs make the map unnecessarily complex, as it requires more resources to navigate, and extracting reliable semantic information becomes challenging. 

In the area of grid-based maps, \cite{9564514} used a straight skeleton to generate a Hierarchical Topological Map (HTM), which results in similar skeleton maps to the Voronoi-based methods. The walls in the map are treated as polygons, and the polygon vertices are moved inward towards the open space along their normal, until they intersect and form a center line. Compared to the Voronoi approach, HTM graphs, generated using a straight skeleton, are less sensitive to noise and generate less complex topological maps while producing smoother paths that are easier for the robots to follow. However, this is at a significantly higher computational cost, making the method not feasible for online map generation.  

Convolutional Neural networks (CNN) can also be used to create topological maps.
\cite{8968111} used a combination of CNN and a segmentation network to detect doorways in an occupancy grid and then used that information to categorize areas into rooms and corridors, thus creating a semantic map that can be used to generate a topological map. However, their method is intended only for indoor environments.  

In comparison to the previous state-of-the-art in topological mapping, the first contribution of this article is based on the establishment of a novel approach to detect and classify intersections from a 2D grid-based map. This method is used to create a semantic map containing the intersections, the openings between intersections, the pathways, the dead ends, and the pathways leading to unexplored frontiers. The second contribution stems from the fact that the proposed method is designed with built-in filtering, making it more robust against noise in the map and environment, especially when compared to existing solutions. The semantic map is then used to generate a topological map containing global navigation paths for a robot. Finally, we demonstrate the gains in sparsity and computational efficiency of the proposed method compared to available state-of-the-art topological methods. 

The rest of this article is organized as follows. In Section~\ref{sec:problem}, the problem solved by the proposed method is defined. In Section~\ref{sec:Implementation}, we introduce and elaborate on the method to detect intersections, build a semantic map, and create a topological map. In Section~\ref{sec:Comparison}, the proposed method is validated, and a detailed comparison with the Voronoi-based mapping methods is presented. Finally, in Section \ref{sec:conclusion}, future research directions are identified, and concluding remarks are made.  

\begin{figure*}[t]
    \centering
    \includegraphics[width=\linewidth]{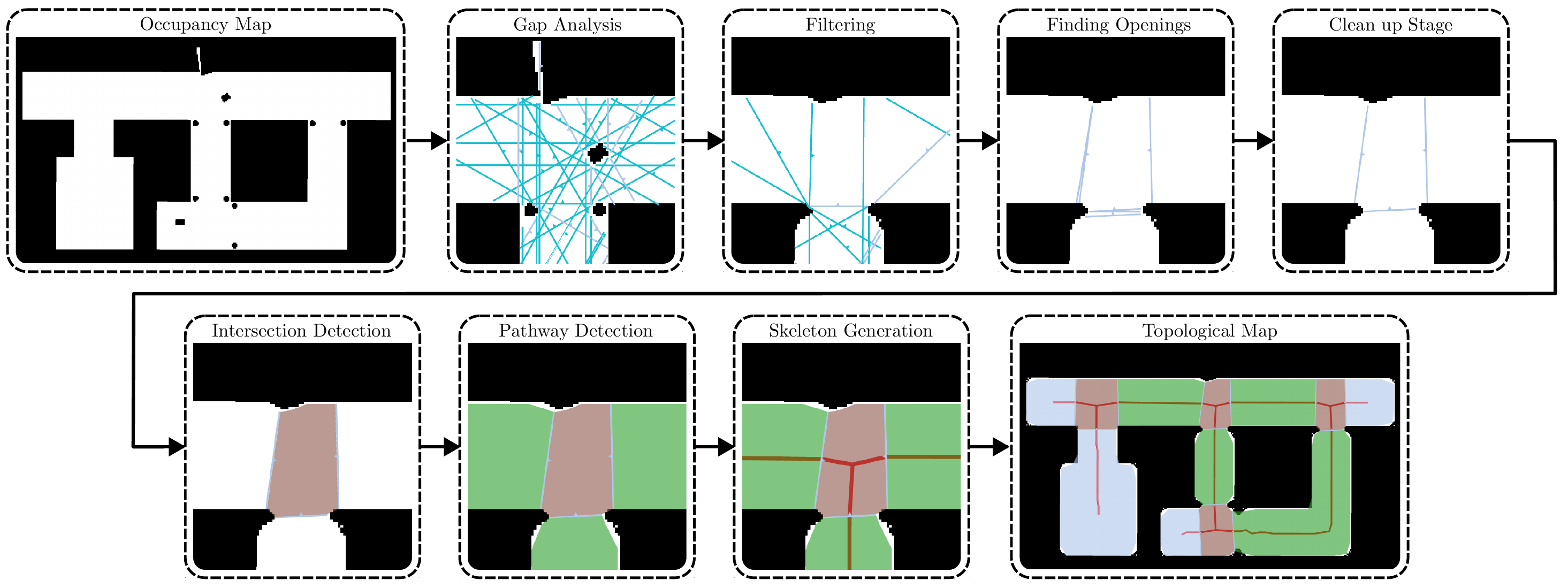}
    \caption{Illustrations of the steps taken by the proposed method to create a topological map. The process takes an occupancy map as input, identifies all intersections, and creates a semantic and topological map. For illustration purposes, the steps are shown on a section of the map, but the method is applied to the complete map.}
    \label{fig:algower}
\end{figure*}
\section{Problem Formulation}
\label{sec:problem}
The semantic and topological mapping solution presented in this article takes as input a given grid based 2D map that is represented as an occupancy grid $M$, where each cell $m_{ij}$, with
$i \in \{0, 1, \ldots, I\}$ and 
$j \in \{0, 1, \ldots, J\}$ can have the value $-1$ for unknown, $0$ for unoccupied, or $1$ for occupied. The occupancy grid has the following matrix representation
\begin{equation}
\label{eq:occogrid}
    M=
    \begin{bmatrix}
        m_{11} & m_{11} & \hdots & m_{1J} \\
        m_{21} & m_{22} & \hdots & \vdots \\
        \vdots & \vdots & \ddots & \vdots \\
        m_{I1} & \hdots & \hdots & m_{IJ}
    \end{bmatrix}.
\end{equation}
The key to deriving a semantic map followed by a topological map is intersection detection, which will be utilized as branch points in the topological map. The overall goal is to identify intersections without preconceptions of the shape, size, or number of connecting rods/corridors. In the sequel, the set of identified intersections $\mathcal{I}$ can then be utilized to find paths between them, dead ends, and pathways leading to unexplored areas of the map. The illustration in Fig. \ref{fig:algower} presents the overall concept of the proposed algorithmic framework while presenting the steps involved in the detection of intersections and the creation of a topological map. These steps will be further analyzed and expanded in the next Section.
\section{Implementation}\label{sec:Implementation}
\subsection{Gap Analysis} \label{sec:gap}
In this subsection, taking the occupancy grid $M$ as an input, we derive the 
gaps present in the map denoted as $g_{ik}$, where $i$ denotes the line of the occupancy grid that the gap belongs to and $k \in \{0, 1, \ldots, K\}$.
The gap $g_{ik}$ is defined as a series of connected unoccupied cells on the same row as represented in Eq. \eqref{eq:gapdef}. The first and last positions of the cells belonging to the gap are denoted by $g_{ikS}$ and $g_{ikE}$ respectively. It should be noted that gaps on the same row cannot overlap or be connected, i.e., $g_{ikS}>g_{i(k-1)E}+1$.
\begin{equation}
\label{eq:gapdef}
    g_{ik}=\{m_{ig_{ikS}}, m_{i(g_{ikS}+1)},m_{i(g_{ikS}+2)}, \hdots m_{ig_{ikE}}\}
\end{equation}
Going ahead, we make the practical assumption that the robot has a finite size $R_{min}$. Then there is a minimum size of the gap that the robot can pass $g_{min}=R_{min}/M_{cell}$, where $M_{cell}$ is the cell size of the occupancy grid. The gaps can belong to one of two groups: gaps that the robot can traverse $g_t$ and non-traversable gaps $g_{nt}$, as defined in Eq. \eqref{eq:gt}.
\begin{equation}
    \label{eq:gt}
    \begin{matrix}
        g_t \: \: = & \{g_{ik}\: |\: g_{ikE}-g_{ikS} \geq g_{min}\}\\
        g_{nt}= & \{g_{ik}\: |\: g_{ikE}-g_{ikS}<g_{min}\}
    \end{matrix}
\end{equation}
Each gap $g_{ik}\in g_t$ is connected to a group of gaps $G_{(i-1)k}$ in the previous line and a group $G_{(i+1)k}$ in the next line, which is defined as in Eq. \eqref{eq:groupGapDef}.
\begin{equation}
\label{eq:groupGapDef}
    \begin{aligned}
        G_{(i\pm 1)k}=\{g_{(i \pm 1)k}\:|&\: g_{(i \pm 1)kS} \in g_t,\\
        & g_{(i \pm 1)kS}<g_{ikE},\: g_{(i \pm 1)kE}>g_{ikS} \}
    \end{aligned}
\end{equation}
All $G_{(i\pm 1)k}$ which contain two or more elements are considered a gap detection and as a potential point of opening into an intersection, as they are points in the map where two or more paths intersect.

A problem with the gap analysis method is that it cannot detect all openings in all scenarios. A good example of one of these situations is a perfect T-intersection perpendicular to the scanning direction, which will not result in any openings detected, as the number of connected gaps for each row is always one. To solve this problem, the occupancy map is scanned $n_{dir}$ times using the method described in this subsection. Each time, the map is rotated by $\pi /n_{dir}$ radian and given as new input to the gap analysis method.
%
%
\subsection{Filtering} \label{sec:filter}
It is not uncommon for occupancy maps to have some unoccupied cells classified as unknown in open spaces, thus creating 'holes' in the map. To allow for some holes in the map, gaps are allowed to contain small groups of maximum $f_{uk}$ number of connected unknown cells, as long as it is not the first or last cell in the gap.

Noise in the environment generates unnecessary gap detections that are not part of an intersection. The solution is to apply an additional condition. For the gap $g_{ik}$ and a $G_{(i\pm 1)k}$ that is part of a gap detection, $g_{ik}$ and at least two of the gaps in $G_{(i\pm 1)k}$ need to be connected to a series of gaps with increasing or decreasing $i$, i.e., $g_{ik}\in S$.  The $S$ is as defined in Eq. \eqref{eq:gapCond} with $g_{dep}$ being the number of minimum connected gaps. 
\begin{equation}
\label{eq:gapCond}
    S= \{g_{ik_1}|g_{(i\pm j)k_j}\in G_{(i\pm (j-1))k_{j-1}}\}_{j=2}^{g_{dep}}
\end{equation}
Gap detections that do not satisfy this condition are discarded as false positives. This step is not necessary for the method to work as the later steps in the method will remove these detections, but removing the detections at this stage will save on the computation required to preforms the steps in subsections \ref{sec:findOp} and \ref{sec:clean}.

The gaps in the groups $g_t$ and $g_{nt}$, defined in Eq. \eqref{eq:gt}, are used to remove holes in the map and remove areas in the map that the robot cannot traverse. This is achieved by setting all $m_{ij} \in g_{ik}$, with $g_{ik}\in g_t$, to unoccupied, thus removing all holes of unknown cells in the map. Similarly, $m_{ij} \in g_{ik}$, with $g_{ik}\in g_{nt}$, are set to occupied if $g_{ik}$ is next to occupied cells. 

Small objects and unoccupied cells classified as occupied cells pose a problem in topological mapping as they lead to unnecessary complexity in the map. To mitigate this problem, each gap $g_{ik} \in G_{(i \pm 1)k}$, where $G_{(i \pm 1)k}$ contains two or more elements, is checked to see if they are connected to an object and then the object is removed if it is under a certain size. This is done by collecting the points along one of the walls that the gap is connected to into a group $P_{L}$. If the first and last element in $P_{L}$ are the same and there are fewer elements in $P_{L}$ than the threshold value $f_{obj}$, the object is removed using a polygon filling algorithm by \cite{10.1145/245.248}. This method is then repeated for the other wall that the gap is connected to. 
After removing the object, the end and start of the gap are updated to fit the new map.
%
%
\subsection{Finding Openings}
\label{sec:findOp}
At each gap in $G_{(i\pm 1)k}$ that is part of a gap detection, there is an opening detection $o_t$ that is defined as a pair of points on the occupancy grid $o_t = \{ m_{i_1j_1}, m_{i_2j_2}\}$. These two points, the start and end points of $o_t$ are denoted by $o_{tS}$ and $o_{tE}$ respectively. The present goal is to represent the opening into the intersection with the line between the opening detection points. The opening's start and end points are placed such that the positive normal of the line is between the points $o_{tS}$ and $o_{tE}$ points into the intersection. The aim is to find the $o_{tS}$ and $o_{tE}$ that best describes the openings into intersections. 

This research will assume that the $o_{t}$ that best represents an opening, has its start and end position next to the occupied cells in the occupancy map, and the line between the points does not overlap with the walls. We denote the set of points along the walls that $o_{tS}$ and $o_{tE}$ can connect to by $W_s$ and $W_e$ respectively and the set of all possible configurations $o_{t}$ by $W_o$. Then the optimal opening detection is assumed to be the opening that minimizes the function in Eq. \eqref{eq:openingMin}, where $||o_t||$ is the distance between its points, $d_c$ is the distance to the center of the intersection and $d_{w}$ is a constant to determine the importance of $d_c$. 
\begin{equation}
    \min_{W_o}(||o_t||+d_w*d_{c})
    \label{eq:openingMin}
\end{equation}
As the term $d_c$ in Eq. \eqref{eq:openingMin} is unknown, since the intersection is not yet detected, the method that it is introduced to find the optimal opening focuses on finding a point close to the local minimum value of $||o_t||$ that is close to the gap detection. The initial value of $o_t$ is the start and end position of the gap, $g_{ik} \in G_{ik}$, that is part of a gap detection.
The local minimum is found by keeping the $o_{tE}$ fixed and finding the point in $W_s$ that gives the smallest $||o_t||$, following which $o_{tS}$ is moved to this position. The process is then repeated with $o_{tS}$ fixed and searching over $W_E$. The application of this method will not find the smallest possible $||o_t||$, but it will result in an opening detection that is closer to the gap detection zone, and this is done with the assumption that it will result in minimizing $d_c$ as well. 
\subsection{Clean up Stage}
\label{sec:clean}
\begin{figure}[htbp]
    \centering
    \begin{subfigure}[b]{0.24\textwidth}
         \centering
         \includegraphics[width=\textwidth]{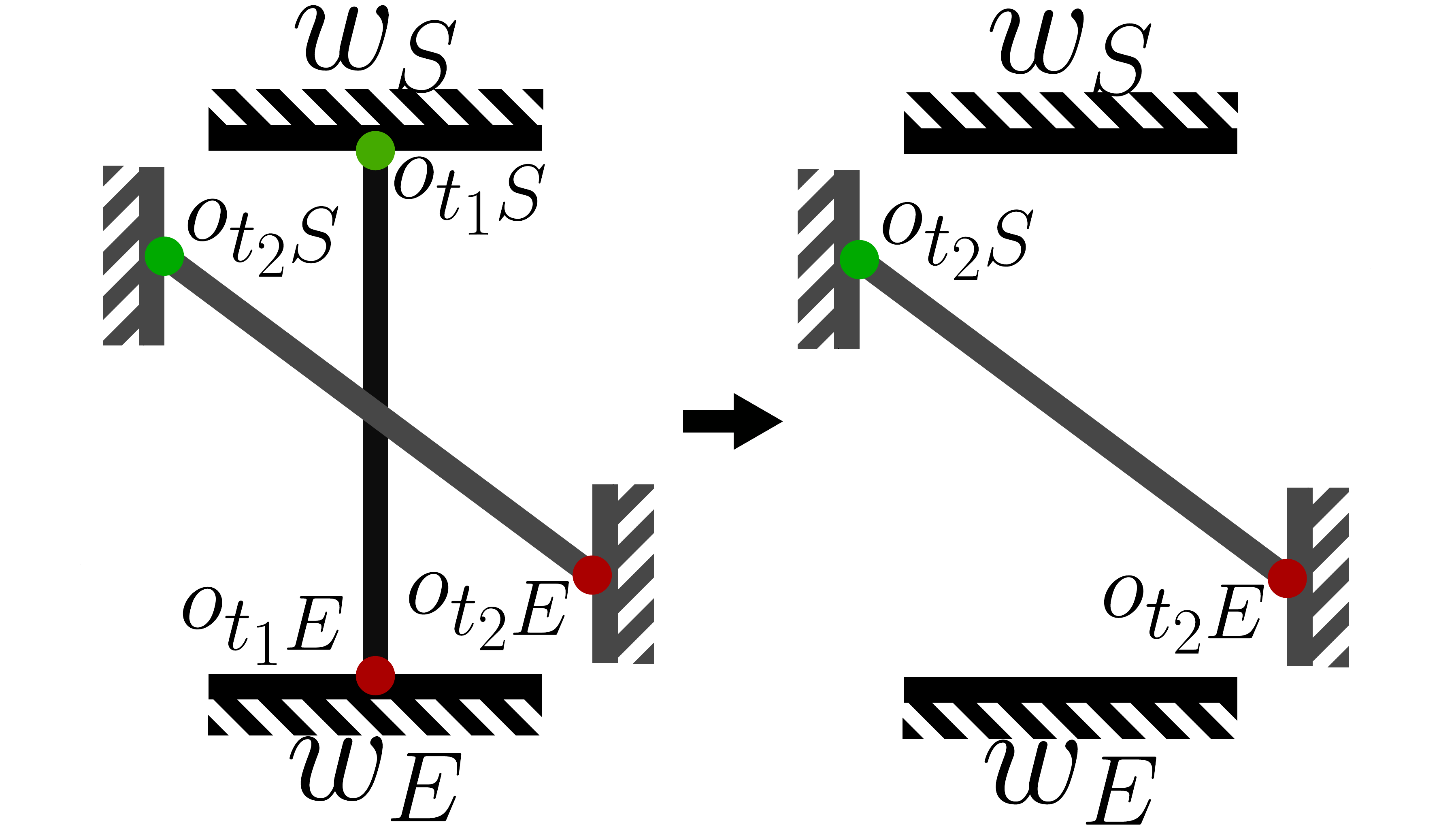}
         \caption{}
         \label{fig:overlapA}
     \end{subfigure}
     \hfill
     \begin{subfigure}[b]{0.24\textwidth}
         \centering
         \includegraphics[width=\textwidth]{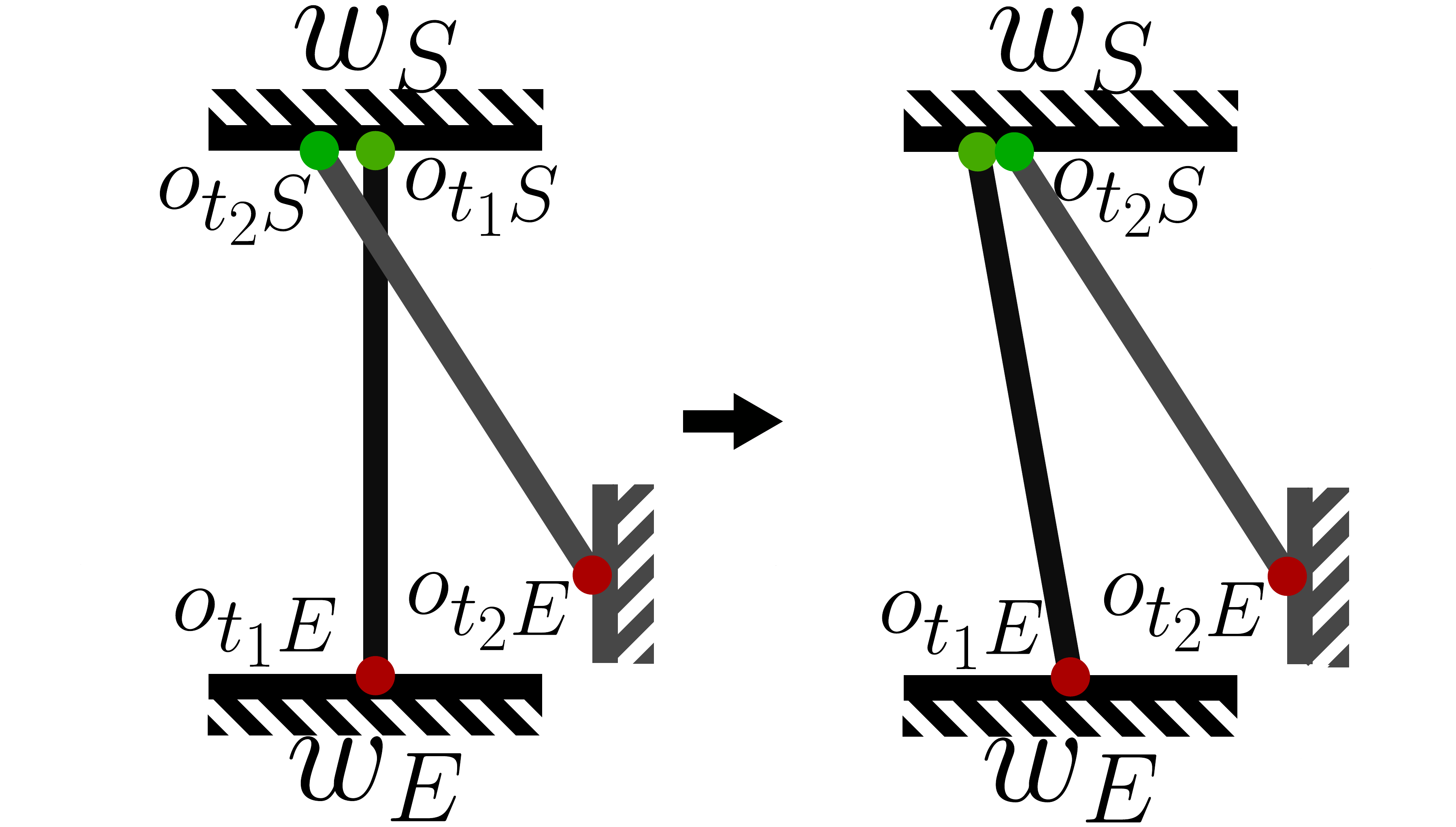}
         \caption{}
         \label{fig:overlapB}
     \end{subfigure}
     \begin{subfigure}[b]{0.24\textwidth}
         \centering
         \includegraphics[width=\textwidth]{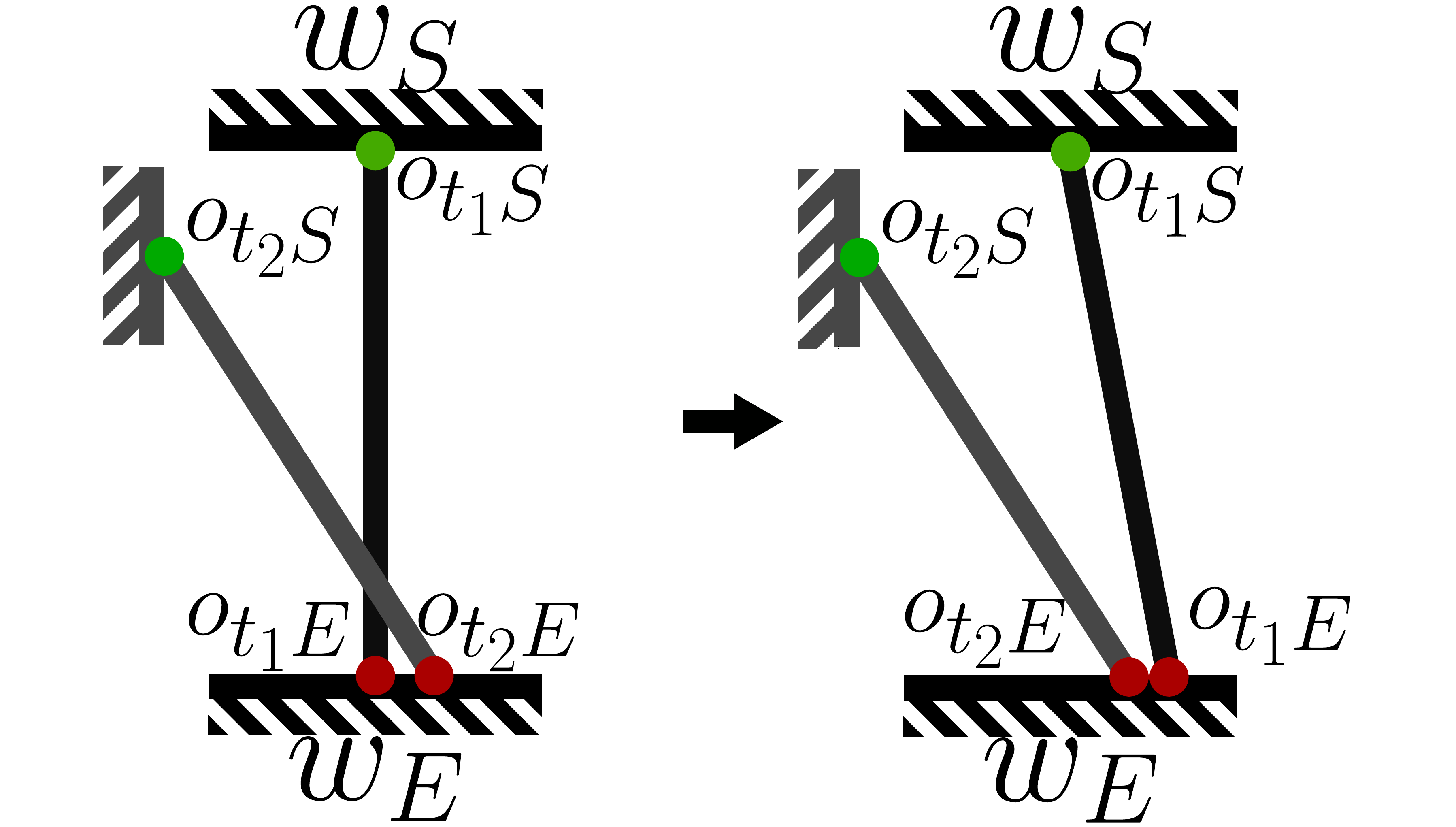}
         \caption{}
         \label{fig:overlapC}
     \end{subfigure}
     \begin{subfigure}[b]{0.24\textwidth}
         \centering
         \includegraphics[width=\textwidth]{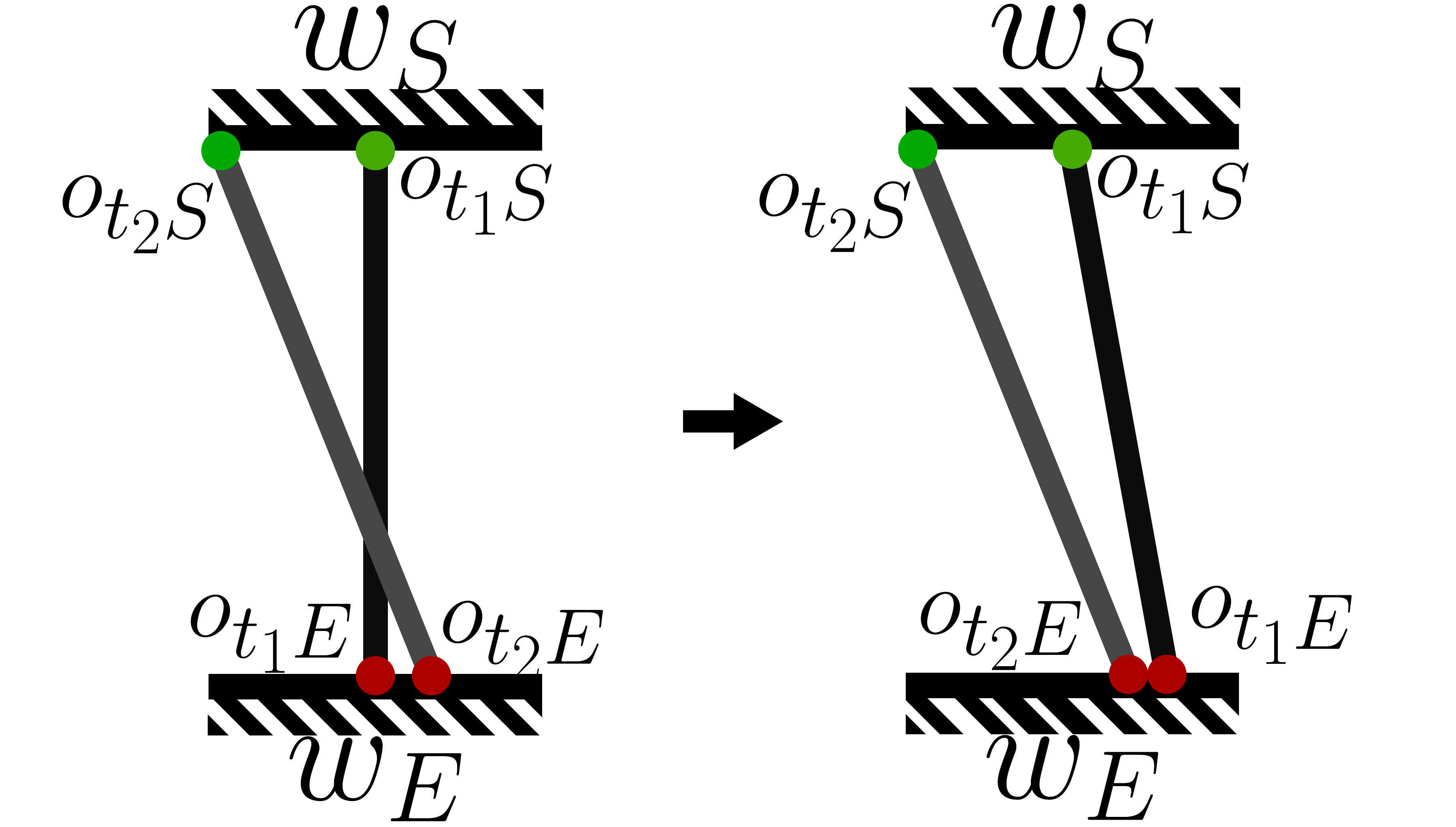}
         \caption{}
         \label{fig:overlapD}
     \end{subfigure}
    \caption{The different kinds of overlapping scenarios and their solutions.}
    \label{fig:overlap}
\end{figure}

\begin{figure*}
    \centering
    \begin{subfigure}[b]{0.24\textwidth}
         \centering
         \includegraphics[width=\textwidth]{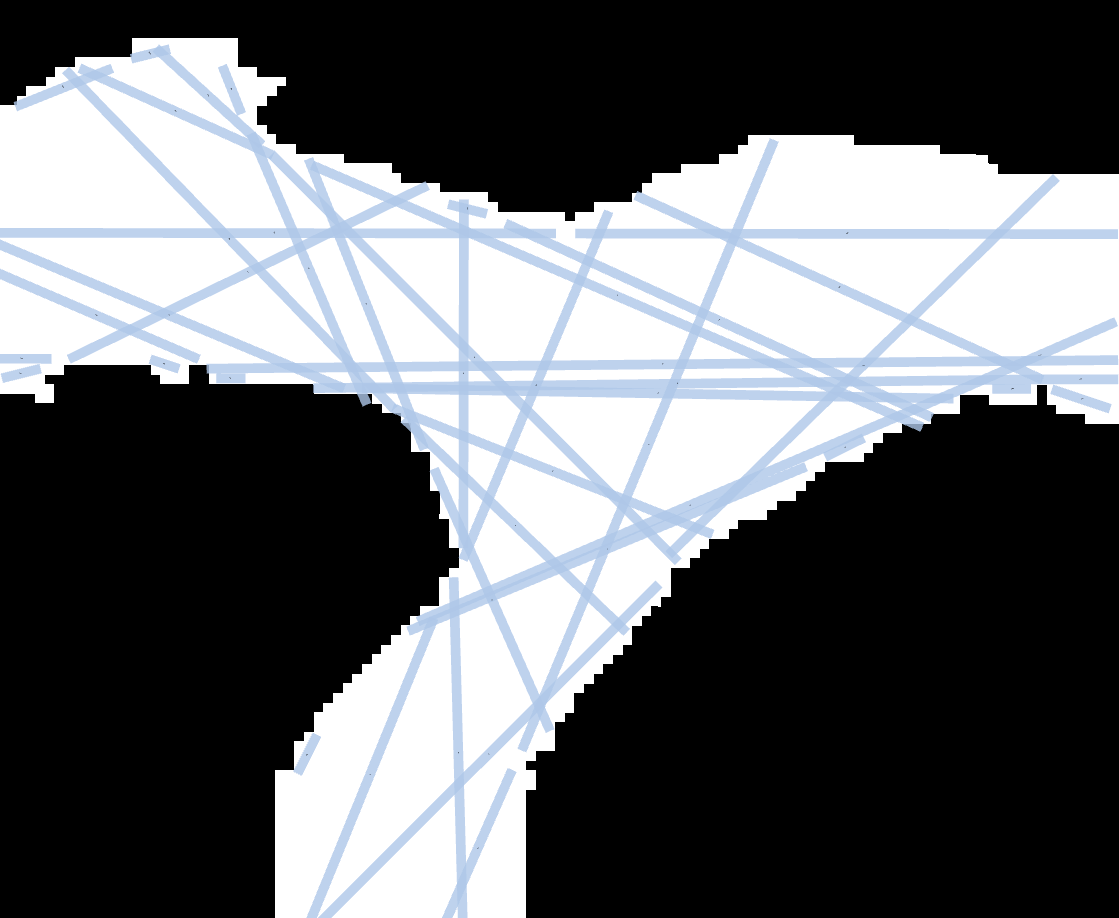}
         \caption{}
         \label{fig:processA}
     \end{subfigure}
     \hfill
     \begin{subfigure}[b]{0.24\textwidth}
         \centering
         \includegraphics[width=\textwidth]{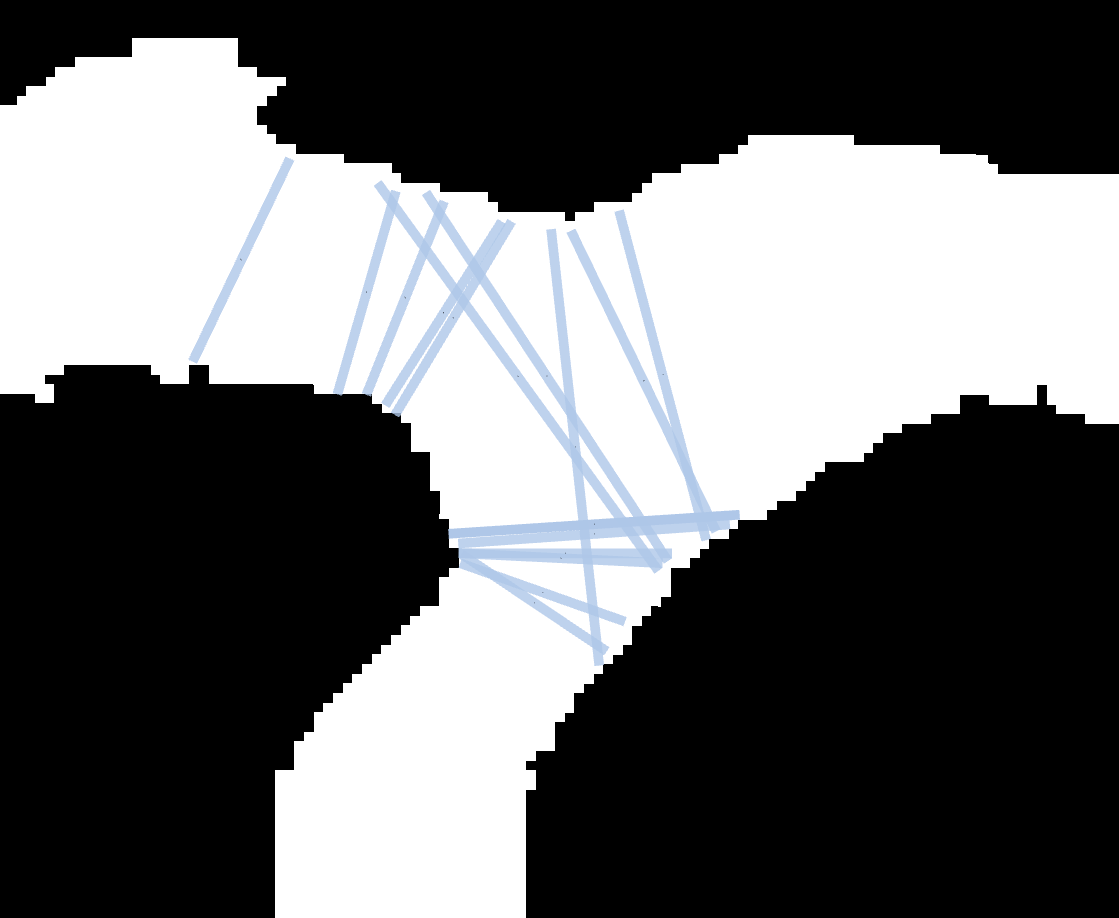}
         \caption{}
         \label{fig:processB}
     \end{subfigure}
     \begin{subfigure}[b]{0.24\textwidth}
         \centering
         \includegraphics[width=\textwidth]{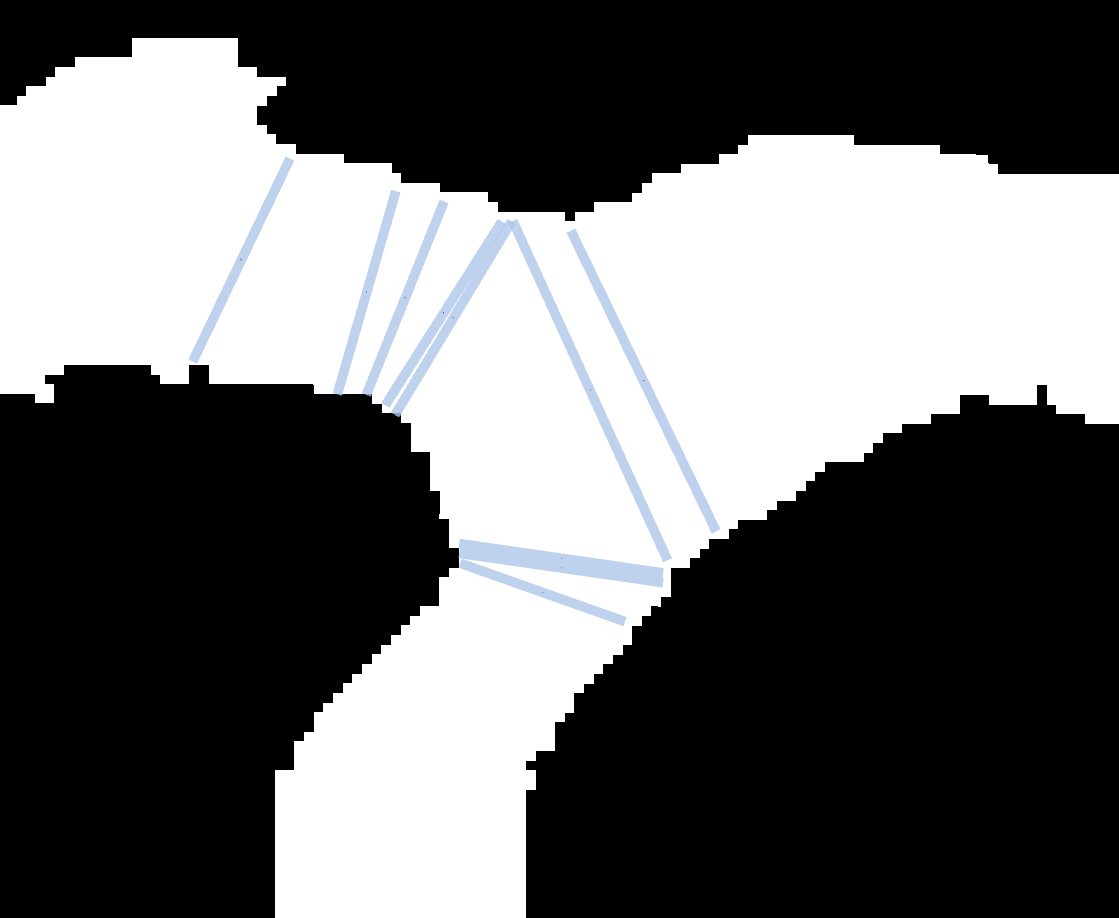}
         \caption{}
         \label{fig:processC}
     \end{subfigure}
     \hfill
     \begin{subfigure}[b]{0.24\textwidth}
         \centering
         \includegraphics[width=\textwidth]{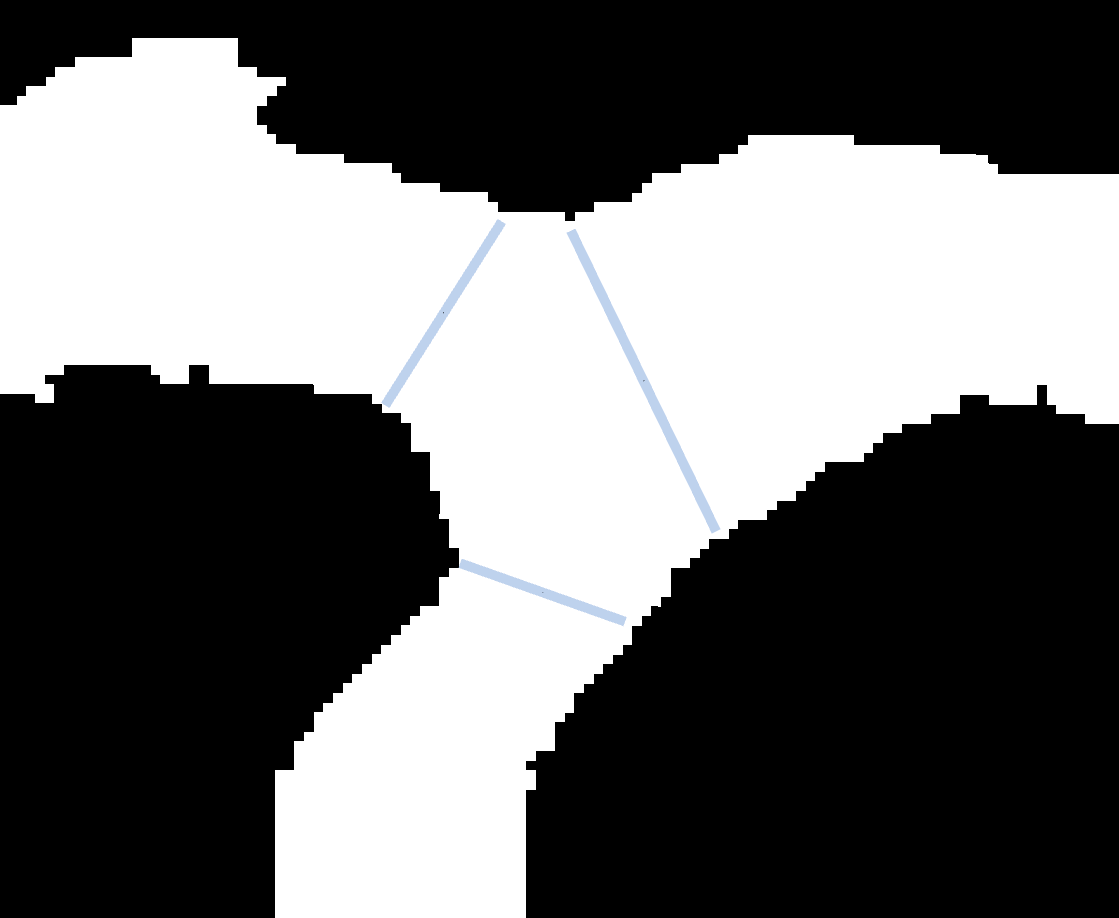}
         \caption{}
         \label{fig:processD}
     \end{subfigure}
    \caption{(a) Initial gap detection. (b) Opening detections were found using gap detection. (c) The opening detections after they have been moved to remove overlaps. (d) The remaining opening detections after unnecessary detection has been removed.}
    \label{fig:process}
\end{figure*}

Each opening detection is independently performed, which results in multiple opening detections for the same opening in the intersection. This also causes overlapping openings, as seen in Fig. \ref{fig:processB} that cause uncertainty about which area belongs to an intersection and therefore it is necessary to ensure that opening detections do not overlap. To solve this problem, when two overlapping openings $o_{t_1}$ and $o_{t_2}$ are detected, the points alongside the wall that $o_{t_1}$ is connected to are collected in both directions in $s_{o}$ steps. The points along the wall are collected as $w_S$ and $w_E$ as defined in Eq. \eqref{eq:walldef}, where $w_{o_{tS}}=o_{t_1S}$, $w_{o_{tE}}=o_{t_1E}$, and $H=s_{o}$. 
\begin{equation}
\begin{matrix}
    w_S=\{ w_{o_{tS}-H},..., w_{o_{tS}-1},w_{o_{tS}},w_{o_{tS}+1},...,w_{o_{tS}+H}\} \\
    w_E=\{ w_{o_{tE}-H},..., w_{o_{tE}-1},w_{o_{tE}},w_{o_{tE}+1},...,w_{o_{tE}+H} \}
\end{matrix}
\label{eq:walldef}
\end{equation}
Depending on how the points in $o_{t_2}$ relate to $w_S$ and $w_E$, different solutions are used, as shown in Fig. \ref{fig:overlap} and with the following corresponding scenarios. 
\begin{itemize}
    \item Scenario (a): $o_{t_2} \not\subset \{w_S, w_E\}$. The shortest opening detection is kept while the other is discarded, as shown in Fig. \ref{fig:overlapA}.
    \item Scenario (b): $o_{t_2} \subset w_S$. If $o_{t_2} \ni m_{ij}=w_{o_{tS}\pm h}$ then $o_{t_1}$ is changed so that $o_{t_1S}=w_{o_{tS}\pm h \pm g}$, where $g$ is the minimum value that doesn't result in an overlap between $o_{t_1}$ and $o_{t_2}$, as shown in Fig. \ref{fig:overlapB}.
    \item Scenario (c): $o_{t_2} \subset w_E$. If $o_{t_2} \ni m_{ij}=w_{o_{tE}\pm h}$ then $o_{t_1}$ is changed so that $o_{t_1E}=w_{o_{tE}\pm h \pm g}$, where $g$ is the minimum value that doesn't result in an overlap between $o_{t_1}$ and $o_{t_2}$, as shown in Fig. \ref{fig:overlapC}.
    \item Scenario (d): $o_{t_2} \subset w_S$, $o_{t_2} \subset w_E$, $o_{t_2}\ni m_{i_1j_1}=w_{o_{tS}\pm h_1}$ and $o_{t_2}\ni m_{i_2j_2}=w_{o_{tE}\pm h_2}$. If $h_1<h_2$ then $o_{t_1}$ is changed so that $o_{t_1S}=w_{o_{tS}\pm h_1 \pm g}$ else $o_{t_1E}=w_{o_{tE}\pm h_2 \pm g}$, where $g$ is the minimum value that doesn't result in an overlap between $o_{t_1}$ and $o_{t_2}$, as shown in Fig. \ref{fig:overlapD}.
\end{itemize}
It is not uncommon to have multiple opening detection in the same opening, as shown in Fig. \ref{fig:processC}. Thus, the aim is to remove all opening detections, except the one that best satisfies the criteria stated in Eq. \eqref{eq:openingMin}. In this case, an approach that is similar to the one proposed for resolving overlapping is used, where firstly, the points along the walls connected to $o_{t_1}$ are collected as $w_S$ and $w_E$ as defined in Eq. \eqref{eq:walldef}, where $H=s_{c}$. With increasing $h$ in $w_{o_{tS}+h}$, the wall is followed counter-clockwise and in $w_{o_{tE}+h}$, the wall is followed clockwise from the perspective of the center of $o_{t_1}$. 

If there exists a  $o_{t_2}$ such that $o_{t_2} \subset w_S$, $o_{t_2} \subset w_E$, $o_{t_2}\ni m_{i_1j_1}=w_{o_{tS}\pm h_1}$ and $o_{t_2}\ni m_{i_2j_2}=w_{o_{tE}\pm h_2}$, then it is assumed that $o_{t_2}$ is in the same opening as $o_{t_1}$. To determine which opening is to be removed, the condition in Eq. \eqref{eq:celning_test} is used.   
\begin{equation}
    \label{eq:celning_test}
    ||o_{t_1}||+ d_{w}*\frac{h_1+h_2}{2} < ||o_{t_2}||
\end{equation}
If the condition is true, $o_{t_2}$ is removed, else $o_{t_1}$ is removed. 
\subsection{Detecting intersections}
\label{sec:detectInter}
Intersections are areas enclosed by opening detections. Identification of an intersection begins by considering an opening detection $o_{t1}$. One of the walls connected to $o_{t1}$ is followed in the direction of the intersection to search for another opening detection $o_{t2}$. As all the opening detections have a direction, the connection between the opening detection should be between their start-to-end points ($o_{t1S} \to o_{t2E}$) or end-to-start points ($o_{t1S} \to o_{t2E}$). The process is then repeated with a new opening detection until a connection between the last opening detection and the first is found, thus closing the loop. The area of the intersections is the polygon created by its opening detections and the points along the walls connecting these opening detections. 

In some situations a bad connection is made, for instance, $o_{t1S} \to o_{t2S}$ and $o_{t1E} \to o_{t2E}$ or a frontier is passed when making a connection. These bad connections imply that an opening detection is missing in the intersection. When a missing opening detection is detected in an intersection, first, all the remaining opening detections are found by searching the other direction in the intersection, $o_{t1S} \to o_{t2E}$ instead of $o_{t1E} \to o_{t2S}$ or vice-versa. Then the missing opening detection is created at the start and end point of the last and first opening detection in the intersection. The new opening detections are then fitted to the opening using the method described in subsection \ref{sec:findOp}. This method allows for up to one opening detection missing in an intersection. 

In some cases, the opening detection can be placed in the wrong order leading to a small intersection with only two openings. This problem is resolved by flipping the opening detections in intersections with only two openings. Flipping an opening detection is done by switching the positions of the end and start points of the opening detection. In case an intersection contains only one opening, the intersection is ignored. 
%
%
\subsection{Finding Paths Between Intersections}
\label{sec:findPath}
A method similar to the one in subsection \ref{sec:detectInter} is used to find the paths between intersections. However, instead of following the walls into the intersection, they are followed away from the intersection. When starting from an opening $o_{t1}$ and using the wall following the approach described in the previous subsection, if a connection $o_{t1S} \to o_{t1E}$ is found, then it is either a dead end or a path to an unexplored area of the map, depending on whether a frontier was passed on the way. Instead, if a connection between two different openings, $o_{t1}$ and $o_{t2}$, is found, it is classified as a path. In cases where a path connects to more than one intersection, it is classified as an intersection. This allows some intersections to be missed while still producing a working topological map. 
On completing the steps described in this subsection, a semantic map will be created.
%
\subsection{Skeleton Generation}
\label{sec:roboPath}
The skeleton representation is generated as a collection of robot paths. The robot paths are generated independently in every path polygon and intersection polygon. In intersections, the robot paths are generated from the center of each opening to the center point of the intersection. In paths between intersections, the robot path is generated between the two openings in the pathway. 

When the path is unobstructed between the start and end points in the occupancy map, the robot path is generated as a straight line between the points. In cases where the path between the start and the end position is obstructed, a local Voronoi graph is generated inside the path polygon or intersection polygon between the points using a thinning algorithm by \cite{10.1145/357994.358023}. The method described in subsection \ref{sec:clean} that is used to remove small objects, is used here to fill in the polygon area, and then the thinning algorithm is utilized on the filled area to approximate the Voronoi graph. The start and end points are locked during the thinning process resulting in a direct path between the two points. All other Voronoi paths generated are discarded. In the cases of dead ends and pathways leading to unexplored areas, a Voronoi graph is created in that area. All Voronoi paths except the longest one connected to the opening of the dead end are discarded. At this point, a complete topological map of the environment has been created and represented as a skeleton map.
\section{Validation and Comparison} \label{sec:Comparison}
\subsection{Validation} \label{sec:validation}
%
The proposed method (PM) was validated on an occupancy grid built from LiDAR data from exploration of the hospital gazebo world\footnote{AWS RoboMaker Hospital World ROS package: https://github.com/aws-robotics/aws-robomaker-hospital-world}. This is an interior environment with rooms, corridors, and intersections.  

\begin{figure*}
    \centering
    \begin{subfigure}[b]{0.245\textwidth}
         \centering
         \includegraphics[width=\textwidth]{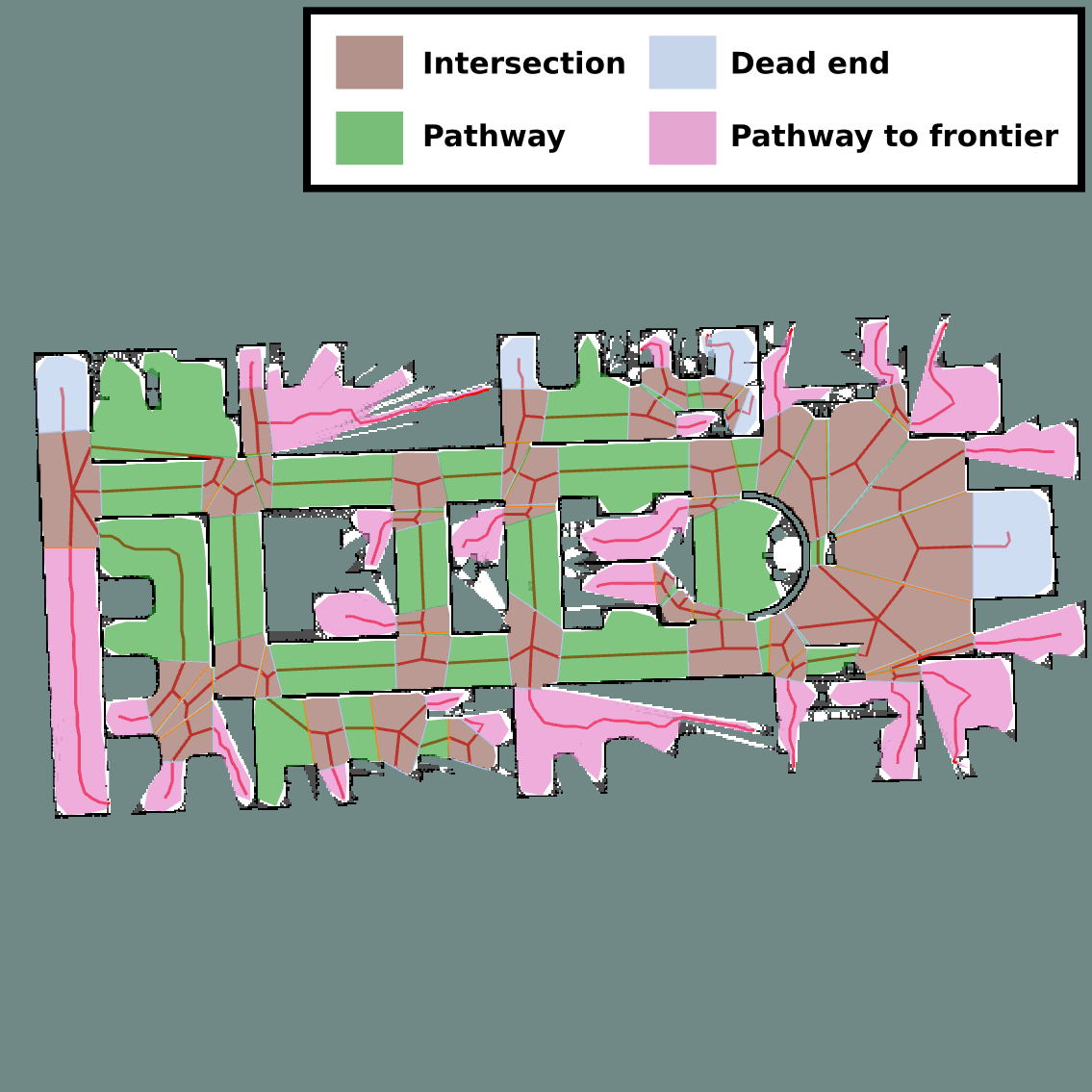}
         \caption{}
         \label{fig:hospitalA}
     \end{subfigure}
     \hfill
     \begin{subfigure}[b]{0.245\textwidth}
         \centering
         \includegraphics[width=\textwidth]{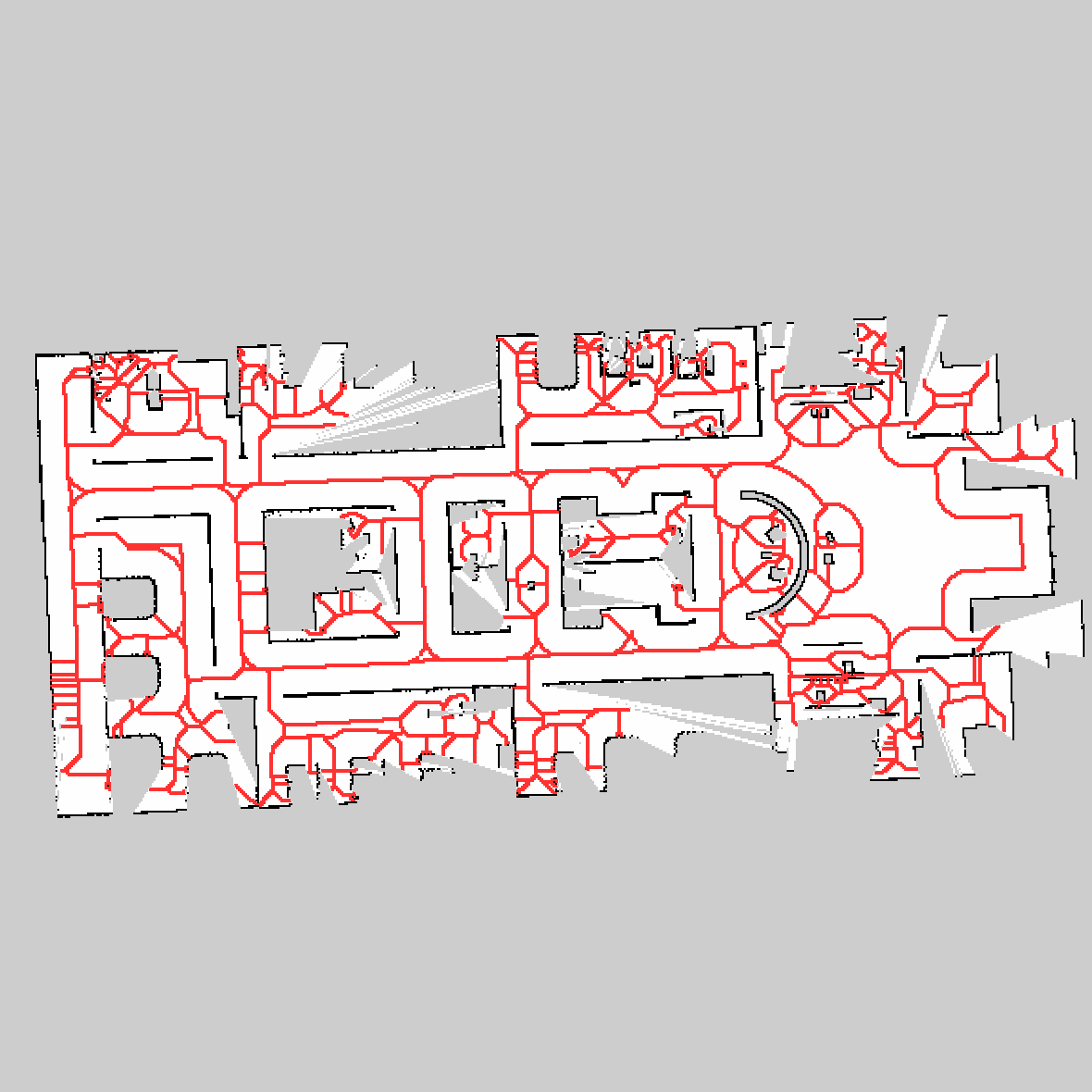}
         \caption{}
         \label{fig:hospitaldB}
     \end{subfigure}
     \begin{subfigure}[b]{0.245\textwidth}
         \centering
         \includegraphics[width=\textwidth]{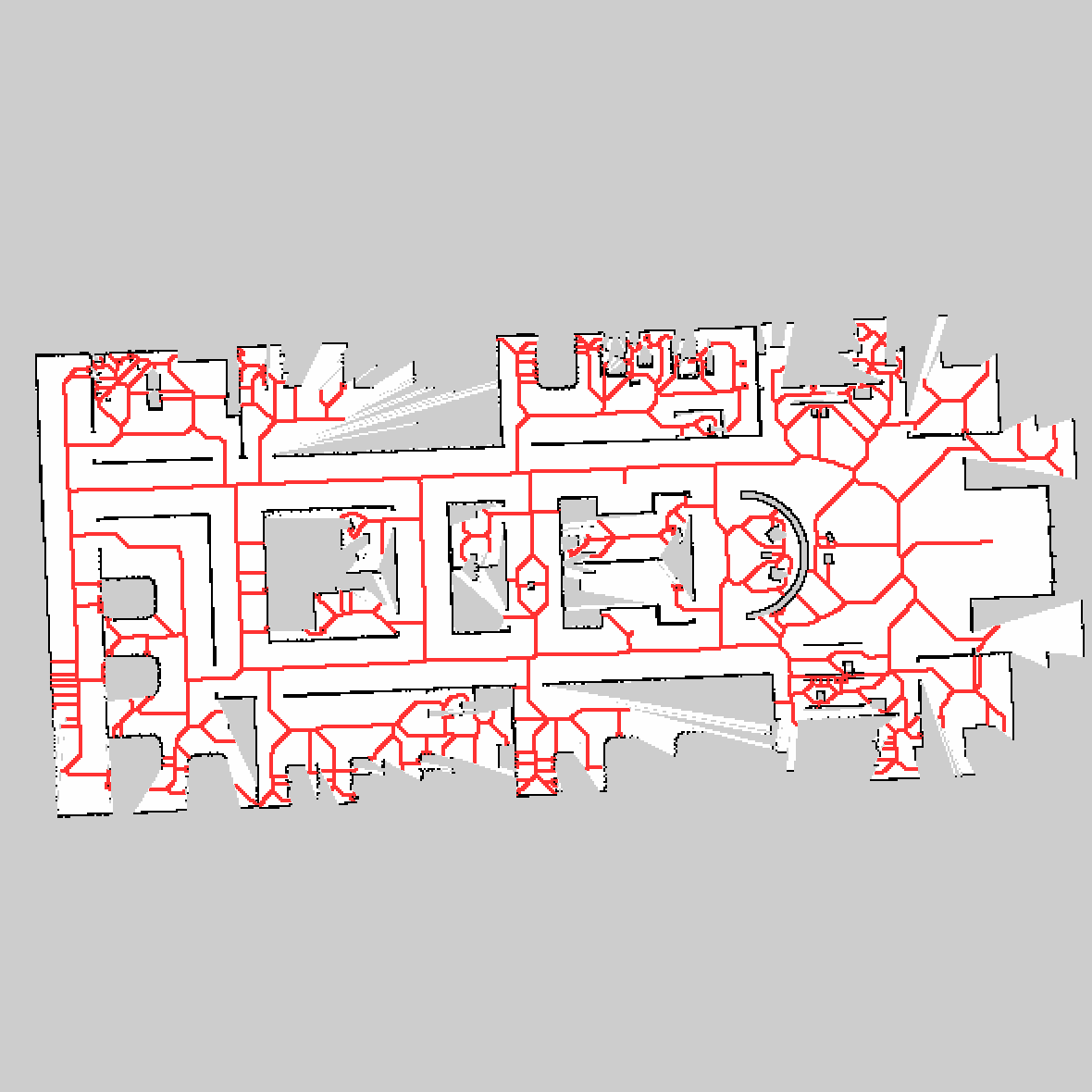}
         \caption{}
         \label{fig:hospitaldC}
     \end{subfigure}
     \begin{subfigure}[b]{0.245\textwidth}
         \centering
         \includegraphics[width=\textwidth]{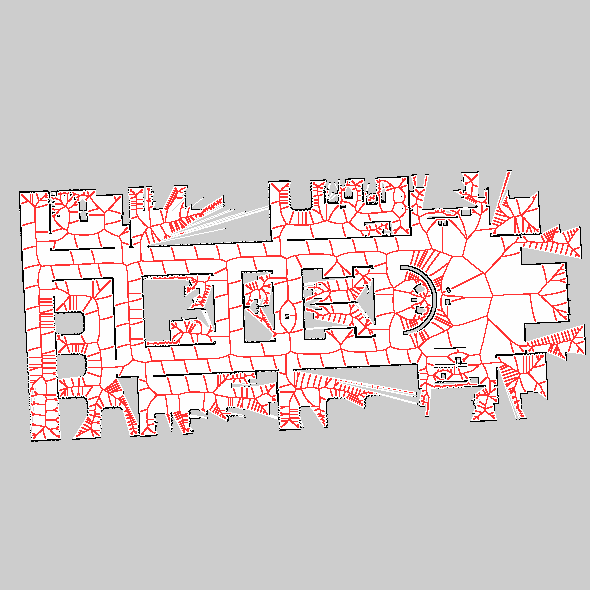}
         \caption{}
         \label{fig:hospitaldD}
     \end{subfigure}
    \caption{Comparison of robot paths generated with different methods in the Hospital map. (a) the result of PM is shown on the filtered map along with the semantically meaningful areas. (b) EVG with a sensor horizon of $1.5m$. (c) RGVG. (d) GVG.}
    \label{fig:hospital}
\end{figure*}

The PM was validated using two different metrics. First, the number of nodes refers to the sum of the number of branch points and endpoints. The other metric is the computation time to process a given occupancy grid and produce a topological map.
In Table \ref{tab:val}, we list the parameters used for the PM. 
\begin{table}[h]
\captionsetup{width=.9\linewidth}
\caption[width=\linewidth]{Values used for the PM in the validation and comparison}
\label{tab:val}
\centering
\begin{tabular}{ll|ll}
\hline
Variable & Value & Variable & Value   \\ \hline
 $g_{min}$& 6 cells & $f_{obj}$ & 40 cells    \\
 $n_{dir}$ & 6 & $d_{w}$ & 0.5 \\
 $f_{uk}$& 1 cell & $s_{o}$ & 600  \\
 $g_{dep}$ & 5 & $s_{c}$ & 50 \\ \hline
\end{tabular}
\end{table}
%
%
In the test, the robot has a minimum working area of $0.6$m. In Table \ref{tab:res}, and Fig. \ref{fig:hospitalA}, we present the validation of the proposed method. In the topological map created by the PM, all major intersections were detected. However, some of the smaller intersections were missed in some of the rooms on the map. 
A problem with the skeleton map generated by the PM is that the robot path does not take into account the size of the robot and can sometimes generate paths very close to the wall, as can be seen in the room in the upper left corner of the 
map in Fig. \ref{fig:hospitalA}. This could be mitigated by utilizing the approach presented in subsection \ref{sec:roboPath}, and more specifically, when deciding whether to generate a local Voronoi or not, instead of checking the collision with a single ray, a rectangle with the same width as the robots could be used to test for a potential collision.

\subsection{Comparison with the state-of-the-art}
The topological map created by PM is compared against GVG (general Voronoi graph \citet{5650794}), RGVG (the reduced general Voronoi graph (\cite{928558})), and the EVG (extended Voronoi graph (\cite{1570793})). The different topological mapping methods were tested on the same occupancy map as in subsection \ref{sec:validation}
The sensor horizon used for the EVG is $1.5$m and the parameters used for the PM are displayed in Table \ref{tab:val}. All maps were generated using Linux, kernel 5.18.6, running on an AMD 5850u CPU.

\begin{table}[h]
\captionsetup{width=.9\linewidth}
\caption{Performance of PM and different state-of-the-art topological mapping methods}
\label{tab:res}
\centering
\begin{tabular}{cccc}
\hline
Method &    Number of Nodes & Computation time (s)\\ \hline
 PM  &   78 & 0.0616  \\
 EVG  &   850 & 0.0513  \\
 RGVG  &   706 & 0.0838  \\      
 GVG  &   2823 & 0.0163  \\  
\end{tabular}
\end{table}

In Fig. \ref{fig:hospital} we show the skeleton maps generated by the different methods. All topology maps are shown over the original map, except for the PM, which is shown over the filtered version of the original map that is created in subsection \ref{sec:filter} and the semantic map generated in subsections \ref{sec:gap} to \ref{sec:findPath}.
As seen in Table \ref{tab:res}, the result indicates that the PM is performing at a similar speed to RGVG and EVG methods while generating a topological map with significantly fewer nodes compared to all of the Voronoi-based solutions. Compared with the RGVG method, which was second in terms of the least amount of nodes generated, the PM produced a map with $89\%$ fewer nodes. 
%
\section{Conclusions}
\label{sec:conclusion}
In this article, a novel method for intersection detection is proposed, which is then used to generate semantic and topological maps. The approach is general as it is designed with no preconceptions on the properties of the map, while taking practical aspects such as the size of the robot into account. The distinguishing feature of this approach is that the topological map is built on top of semantic information, unlike the classical Voronoi-based approaches. The proposed method is shown to generate significantly sparser topological maps, at a rate that is faster than some of the state-of-the-art solutions, which makes it highly suitable for online global navigation of mobile robots. In the future, we plan to develop and optimize the method further and implement it in online scenarios. 
%
\bibliography{bliography}             

\end{document}